\documentclass[11pt]{article}

\usepackage{acl}

\makeatletter
\ifdefined\IfFormatAtLeastTF
   \IfFormatAtLeastTF{2025-06-01}{
      \RemoveFromHook{build/column/before}[lineno]
      \AtBeginDocument{\let\@LN@makecol\@makecol}
      \AddToHook{build/column/after}{%
         \setbox\@outputbox \vbox to\ht\@outputbox{%
            \boxmaxdepth \@maxdepth
            \protected@write\@auxout{}{%
               \string\def\string\@LN@column{\if@firstcolumn1\else2\fi}%
            }%
            \unvbox\@outputbox
         }%
      }
   }{}
\fi
\makeatother

\usepackage{times}
\usepackage{latexsym}
\usepackage[T1]{fontenc}
\usepackage[utf8]{inputenc}
\usepackage{microtype}
\usepackage{inconsolata}
\usepackage{graphicx}
\usepackage{amsmath, amssymb, amsthm}
\usepackage{booktabs}
\usepackage{amsfonts}
\usepackage{dsfont}
\usepackage{algorithm}
\usepackage{algpseudocode}
\usepackage{orcidlink}

\title{Loom: Weaving Diagnostic Strands into Free-Text Consensus \\ via Embedding-Space Reweighting}

\author{
  Ron Begleiter\,\orcidlink{0009-0008-6599-4196},
  Katya Egert Berg\,\orcidlink{0000-0001-6152-1932},
  Gilad Saban\,\orcidlink{0009-0007-1934-5681},
  Gil Shabat\,\orcidlink{0000-0002-7798-3194} \\
  NVIDIA \\
  Tel Aviv, Israel \\
  \texttt{\{rbegleiter, kegertberg, gsaban, gshabat\}@nvidia.com}
}

\begin{document}
\maketitle

\begin{abstract}
Aggregating noisy, conflicting textual hypotheses into a reliable consensus is a fundamental challenge when deploying NLP systems in real-world industrial settings. While monolithic Large Language Model (LLM) agents offer unbounded expressivity for tasks like Root Cause Analysis (RCA), they suffer from context limits, compounding hallucinations, and prohibitive inference latency. Traditional weak supervision offers statistical rigor but is mathematically restricted to discrete classes. We present Loom, a generative consensus framework deployed for real-world RCA that bridges these paradigms. Loom aggregates open-form hypotheses emitted by modular heuristics (diagnostic templates dynamically populated with episode-specific entities, times, and metrics) by projecting them into a continuous embedding space, and resolves conflicting signals with an iterative centroid-based reweighting algorithm. The resulting consensus weights ground a single lightweight LLM synthesis step. Evaluated on the OpenRCA benchmark, Loom occupies the accuracy--efficiency Pareto frontier: it matches a state-of-the-art autonomous agent on Bank and Market-2 and trails on Market-1 and Telecom, while using a single LLM call per incident on all four datasets ($\sim$26$\times$ faster; $\sim$33$\times$ with an 8B-parameter synthesizer). We discuss our deployment experience, highlighting lessons learned regarding the trade-offs between agentic depth and inference latency, negative results in redundancy detection, and how deterministic consensus fosters trust among Subject Matter Experts~(SMEs).
\end{abstract}

\section{Introduction}

The ability to aggregate diverse, noisy textual signals into a cohesive consensus is a central challenge in natural language processing. As NLP systems are increasingly deployed to analyze massive, multimodal data streams in industrial settings, the need for robust semantic aggregation has outpaced traditional methodologies. In the domain of automated Root Cause Analysis (RCA) for large-scale computing environments, diagnosing failures requires generating statistically consistent, highly specific semantic analyses from diverse sources such as logs and telemetry. 

Deploying monolithic Large Language Models (LLMs) as autonomous agents \citep{xu2025openrca} offers unbounded expressivity but introduces severe practical limitations: context window exhaustion, compounding hallucinations, and prohibitive inference latency and cost. Conversely, weak supervision frameworks \citep{ratner2017snorkel} excel at denoising heuristic rules but are mathematically restricted to discrete, predefined categorical classes. They are structurally incapable of aggregating unconstrained, free-text semantic descriptions.

The core problem lies in bridging these two paradigms: how can we mathematically aggregate \textit{open-form, episode-specific} textual hypotheses without relying entirely on the fragile, unconstrained reasoning and high latency of an iterative LLM loop?

To address this gap, we present \textit{Loom}, a continuous-space generative consensus framework designed for real-world deployment. Loom decomposes complex reasoning tasks into modular, programmatic heuristics (Diagnostic Strands). Rather than voting on discrete labels, these strands emit templated hypotheses whose slots are filled from telemetry. Loom aggregates these open-form outputs by projecting them into a continuous vector space and applying an iterative centroid-based reweighting algorithm. Finally, a synthesis LLM uses this mathematically denoised, ordered evidence to generate a single, coherent narrative.

Crucially, Loom addresses key industrial constraints: it provides an auditable, deterministic consensus mechanism that fosters trust among Subject Matter Experts (SMEs), and it operates at a fraction of the latency and cost of iterative agentic loops.

\paragraph{Contributions.} Our contributions are: (i) a generative consensus framework deployed for real-world RCA that extends weak supervision to templated, episode-specific hypotheses; (ii) an iterative centroid-based reweighting algorithm that resolves conflicts deterministically, bypassing costly LLM debate; (iii) real-world case studies demonstrating Loom's deployment in large-scale computing environments; (iv) an OpenRCA evaluation placing Loom on the accuracy--efficiency Pareto frontier, with a $\sim$26--33$\times$ inference speedup over the RCA-Agent baseline on every dataset; and (v) a discussion of negative results and lessons learned from deployment, specifically regarding redundancy detection and the limits of single-shot synthesis.

\section{Background and Related Work}

Troubleshooting failures in large-scale distributed compute systems is notoriously difficult, primarily due to the stochasticity of entities and complex networking protocols. For example, a typical LLM training job spans hundreds or thousands of compute nodes connected via diverse backbones \citep{jiang2024megascale, jiang2025l4}. When a failure occurs, identifying the true culprit is challenging as it is often masked by ``innocent'' components affected by collateral damage. In practice, troubleshooting heavily relies on human Subject Matter Experts (SMEs). Automated solutions must not only be accurate but also fast, cost-effective, and auditable.

\paragraph{AIOps and Automated Log Analysis.} Large-scale infrastructure management relies heavily on automated log analysis, statistical anomaly detection, and comprehensive observability tools (e.g., MegaScale \citep{jiang2024megascale} and L4 \citep{jiang2025l4}). While highly efficient at data extraction and spatial-temporal pattern matching, these methods fundamentally lack generative semantic capabilities. In other words, they are good at locating symptoms; however, they lack the ability to produce a cohesive RCA description without human SME intervention.

\paragraph{Traditional Weak Supervision.} Weak supervision frameworks, such as Data Programming and Snorkel \citep{ratner2016data, ratner2017snorkel}, revolutionized programmatic labeling by aggregating noisy heuristic rules to generate probabilistic training data. Subsequent advancements drastically improved scalability by using covariance matrices \citep{varma2019learning} and triplet methods \citep{fu2020fast} to learn rule dependencies without ground truth. However, these mathematical formulations are structurally bound to discrete, categorical label spaces. They rely on exact-match voting matrices; if two heuristics output slightly different textual explanations for the same underlying bug, traditional weak supervision treats them as completely disagreeing. Consequently, they are entirely incapable of aggregating unconstrained, free-text semantic descriptions.

\paragraph{Autonomous RCA Agents.} Recent approaches deploy foundational LLMs as autonomous, multi-step reasoning (e.g., ReAct) agents to iteratively navigate telemetry and diagnose root causes \citep{xu2025openrca, yao2023react}. While highly expressive, these bottom-up agentic solutions struggle with unbounded search spaces. Iterative LLM loops (e.g., searching, reading, deciding the next query) incur massive token costs and prohibitive inference latency. Furthermore, lengthy reasoning trajectories suffer from compounding hallucinations, where a single poor retrieval step can permanently derail the agent's diagnosis. Their reliance on generic reasoning also leads to contextual conflation, frequently missing subtle but critical technical nuances.

\paragraph{LLM Consensus and Multi-Agent Debate.} Within the NLP community, efforts to resolve conflicting hypotheses have focused on iterative LLM prompting. Techniques like Self-Consistency \citep{wang2023selfconsistency} sample multiple paths and select the majority vote, while Multi-Agent Debate \citep{du2024improving} instantiates multiple LLMs to iteratively critique each other's outputs. While effective for general reasoning, these approaches scale poorly to massive evidence spaces due to prohibitive token costs and latency. Furthermore, while multi-agent debate is designed to mitigate the ``Degeneration-of-Thought'' phenomenon \citep{liang2024encouraging} (where single models struggle to resolve conflicts once anchored to an initial hypothesis), the iterative generation required makes it too slow for real-world deployment. Loom avoids these pitfalls by shifting conflict resolution to a highly efficient, continuous mathematical embedding space.

\paragraph{Retrieval-Augmentation and LLM-as-Judge Ensembles.} RAG \citep{lewis2020rag}, Fusion-in-Decoder \citep{izacard2021leveraging}, and Self-RAG \citep{asai2024selfrag} couple a retriever with a generative reader but leave cross-passage aggregation to the LLM's in-context reasoning. A parallel line uses LLMs as judges or aggregators: MT-Bench \citep{zheng2023judging}, heterogeneous panels \citep{verga2024replacing}, and pairwise-ranking ensembles \citep{jiang2023llmblender}. Loom is complementary: it pre-aggregates structured hypotheses in continuous space, so the synthesis LLM receives a denoised ranked slate rather than raw passages or many completions. Conflicts are resolved deterministically in embedding space, and aggregation cost does not scale with the number of hypotheses or repeated LLM invocations.

\section{The Loom Framework}

\begin{figure*}[t]
\centering
\includegraphics[width=\textwidth]{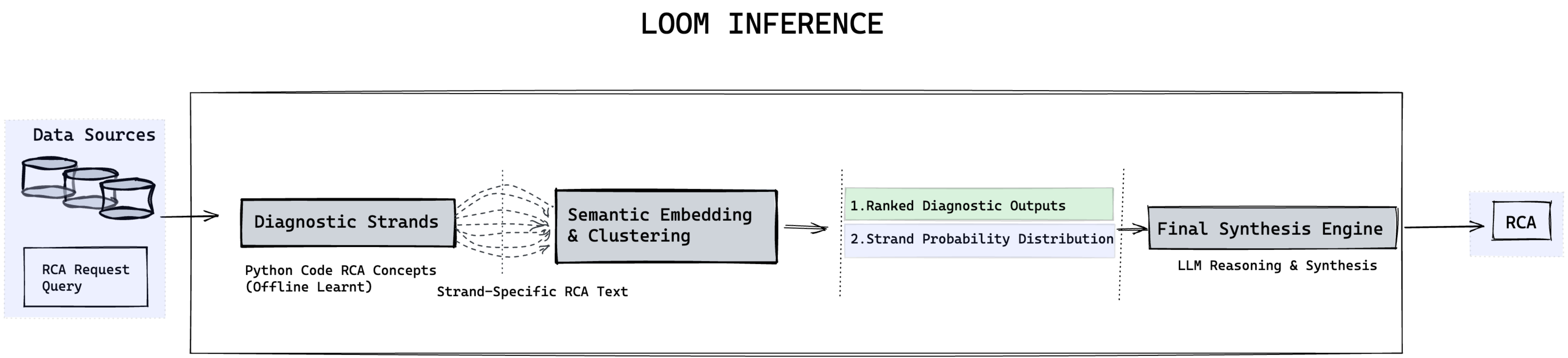}
\caption{The Loom inference pipeline. Diverse data sources are evaluated by programmatic Diagnostic Strands that emit templated, episode-specific Root Cause Analysis (RCA) hypotheses. These outputs undergo semantic embedding and centroid reweighting to identify the most reliable consensus, yielding an importance weighting and similarity map over the strands. Finally, a synthesis LLM reasons over the ranked diagnostic outputs and generates a cohesive final RCA report.}
\label{fig:loom_inference}
\end{figure*}

As illustrated in Figure \ref{fig:loom_inference}, Loom decomposes diagnostic evaluation into modular, programmatic heuristics whose textual outputs are mathematically denoised. At inference time, we feed the entire corpus of heuristics with the data context of a specific incident. Relevant heuristics produce a textual RCA, while irrelevant ones abstain. We compute an importance weighting over the outputs via an iterative embedding-centroid reweighting algorithm, and feed this consensus into a synthesis LLM.

\subsection{Diagnostic Strands (DSs)}
\label{sec:strands}

The foundational units of Loom are Diagnostic Strands (DSs). A DS is a programmatic heuristic (e.g., a Python script) that evaluates structured data. Rather than selecting from a predefined discrete class, a DS emits a diagnostic template whose slots are populated from telemetry (hostnames, KPI names, counts, timestamps), or abstains if its conditions are not met. Appendix~\ref{app:catalog} reports catalog scale, runtime firing, instantiation counts, and authoring cost. Concrete DS listings appear in Appendices~\ref{app:ds_example} and~\ref{app:ib_switch}; Appendix~\ref{app:worked_example} walks through an OpenRCA inference.

For real-world deployment, DSs are either authored by domain experts or automatically extracted via LLMs from historical incident tickets and documentation (see Appendix~\ref{app:offline_learning}). By pre-compiling troubleshooting knowledge into code, Loom avoids the latency and hallucination risks of an agent blindly searching telemetry at runtime.

\subsection{Iterative Embedding-Centroid Reweighting}
\label{sec:reweighting}

Suppose $M$ DSs fire and produce episode-specific templated descriptions. Loom aggregates these using an iterative embedding-centroid reweighting algorithm (Algorithm~\ref{alg:reweighting}), measuring semantic alignment via cosine similarity $\operatorname{sim}(a, b)$.

\paragraph{Step 1: Static Redundancy Detection.} Prior to runtime, Loom computes a similarity matrix over the docstrings of all DSs using an embedding map $\phi$. If the similarity between two DSs exceeds a predefined threshold $\tau$ (a hyper-parameter controlling the strictness of deduplication), they are grouped, and their effective weights are divided by the group size to prevent redundant rules from dominating.

\paragraph{Step 2: Initialization.} Each DS $i$ receives an initial weight $w_i^{(0)}$ based on expert-curated reliability metadata, giving high-reliability heuristics a head start.

\paragraph{Step 3: Dynamic Embedding.} During runtime, the templated outputs of fired DSs are embedded into a continuous vector space, yielding embeddings $e_i = \phi(\text{output}_i) \in \mathbb{R}^d$.

\paragraph{Step 4: Iterative Reweighting.} Starting from $w_i^{(0)}$, Loom alternates two updates. The \textbf{Centroid-step} computes a weighted centroid of the embeddings:
\begin{equation}
    c^{(t)} \;=\; \frac{\sum_{i=1}^M \tilde{w}_i^{(t)}\, e_i}{\sum_{i=1}^M \tilde{w}_i^{(t)}},
\end{equation}
where $\tilde{w}_i^{(t)}$ is adjusted for redundancy. The \textbf{Weight-step} updates each DS's accuracy estimate by its alignment with the centroid:
\begin{equation}
    w_i^{(t+1)} \;=\; \operatorname{sim}\!\big(e_i,\, c^{(t)}\big).
\end{equation}
This process assigns higher weights to consensus-aligned heuristics and down-weights outliers. The entire aggregation runs in milliseconds, contributing negligibly to inference latency.

\begin{algorithm}[t]
\caption{Iterative Embedding-Centroid Reweighting.}
\label{alg:reweighting}
\begin{algorithmic}[1]
\Require Fired DS embeddings $\{e_i\}_{i=1}^M \subset \mathbb{R}^d$; initial reliability weights $\{w_i^{(0)}\}_{i=1}^M$; redundancy groups $\{\mathcal{G}_k\}$ with $g(i)$ denoting the group containing DS $i$; tolerance $\varepsilon > 0$; maximum iterations $K$.
\Ensure Consensus weights $w \in \mathbb{R}^M$.
\State $w_i \gets w_i^{(0)}$ for $i = 1, \dots, M$
\State $t \gets 0$
\Repeat
    \State $\tilde w_i \gets w_i / |\mathcal{G}_{g(i)}|$ \Comment{Static redundancy adjustment}
    \State $c \gets \big(\textstyle\sum_i \tilde w_i\, e_i\big) \big/ \big(\textstyle\sum_i \tilde w_i\big)$ \Comment{Centroid-step}
    \State $w_i^{\text{new}} \gets \operatorname{sim}(e_i,\, c)$ for $i = 1, \dots, M$ \Comment{Weight-step}
    \State $\delta \gets \max_i \big| w_i^{\text{new}} - w_i \big|$
    \State $w \gets w^{\text{new}}$; \quad $t \gets t + 1$
\Until{$\delta < \varepsilon$ \textbf{ or } $t \geq K$}
\State \Return $w$
\end{algorithmic}
\end{algorithm}

\subsection{Textual Resolution and LLM Synthesis}
\label{sec:synthesis}

The reweighting yields an importance weighting that induces a strict ordering over the raw textual outputs. Finally, this ordered evidence is fed into an LLM to synthesize a single, coherent RCA. Because the LLM operates solely on a small, mathematically denoised context window rather than raw telemetry, this step is highly efficient and robust against hallucination. The LLM is strictly prompted to act as a synthesizer, preserving technical details verbatim and prioritizing higher-weighted observations (see Appendix~\ref{app:synthesis_prompt}).

\section{Real-World Deployment and Evaluation}

Loom is designed for environments where ground-truth data is scarce, search spaces are vast, and operational constraints (cost, latency, trust) are strict. To demonstrate its versatility and efficacy, we highlight its application across two distinct domains: NVIDIA production datacenters (where we present qualitative case studies) and the public OpenRCA benchmark (where we present a quantitative empirical evaluation).

\subsection{Deployment in Large-Scale Environments}

We deployed Loom in NVIDIA production datacenters to diagnose both hard distributed job failures and silent networking performance degradations. By shifting the conflict resolution burden to the continuous consensus stage, Loom provides operators with a deterministic, auditable evidence trail, which is a critical requirement for SME adoption. 

\paragraph{Case Study: Distributed Training Job Failure.}
Consider a distributed training job that failed across a 512-node cluster. Traditional log analysis tools flagged hundreds of generic timeout errors, leaving operators to manually sift through the noise. Loom processed the telemetry and produced the following synthesized RCA:

\begin{quote}
\small
\textbf{Title:} Critical fabric link failure initiated by REMAP non-fatal errors on hosts \texttt{node-011-T07} and \texttt{node-011-T15}.

\textbf{Narrative:} The failure cascade began with physical-layer link training instability on hosts \texttt{node-011-T07} and \texttt{node-011-T15}, evidenced by concurrent link-layer retransmissions (480M retries) and elevated raw bit error rates ($7.00\times10^{-6}$). This instability triggered preemptive context removals across the local domain, ultimately causing severe throughput drops on 337 downstream hosts and a switch-correlated collapse pattern affecting 507 hosts total. The root cause is a localized hardware failure (likely a faulty transceiver or cable) within the \texttt{node-011} domain.
\end{quote}

This output demonstrates Loom's ability to pinpoint the true root cause (the specific faulty nodes and hardware layer) while accurately describing the massive blast radius (the 507 downstream ``victim'' hosts) without being distracted by the sheer volume of secondary timeout errors.

\paragraph{Case Study: Silent Performance Degradation.}
Consider a scenario where a distributed training workload experiences a sudden 30\% drop in throughput. No hard failures (e.g., link downs, hardware faults) are reported by the orchestrator. During its periodic snapshot, Loom's autonomous health check evaluates the cluster telemetry. Several DSs fire and are assigned importance weights by the iterative reweighting algorithm:

\begin{itemize}
    \item \texttt{ds\_broken\_dcqcn\_loop} \textbf{(Weight: 0.94):} A composite strand that cross-references switch-side Explicit Congestion Notification (ECN) marking with host-side Congestion Notification Packet (CNP) handling. It fires because switches are marking packets, but hosts are not reacting.
    \item \texttt{ds\_ecn\_marked\_high} \textbf{(Weight: 0.85):} Detects severe congestion at the spine switches (ECN marks $>50$k/s).
    \item \texttt{ds\_cnp\_handled\_low} \textbf{(Weight: 0.82):} Detects that hosts are receiving almost no CNPs ($<10$/s).
    \item \texttt{ds\_rx\_pause\_high} \textbf{(Weight: 0.76):} Detects elevated Priority Flow Control (PFC) pause frames as switch buffers fill up.
\end{itemize}

The synthesis LLM processes this weighted evidence and produces the following RCA:

\begin{quote}
\small
\textbf{Title:} Severe fabric congestion cascade caused by a broken DCQCN feedback loop.

\textbf{Narrative:} The fabric is experiencing severe congestion at the spine layer, evidenced by high ECN marking rates ($>50$k/s) and elevated PFC pause frames. However, host-side telemetry shows almost no CNPs being handled ($<10$/s), indicating that the DCQCN congestion control feedback loop is broken. Because senders are not being throttled appropriately, the network is overwhelmed, leading to buffer exhaustion and the observed drop in effective throughput. The root cause is likely a configuration drift or routing issue preventing CNPs from reaching the sender hosts.
\end{quote}

This example highlights how Loom uses the continuous-space consensus to assign the highest importance to the composite causal strand (\texttt{ds\_broken\_dcqcn\_loop}), allowing the LLM to synthesize a coherent narrative out of seemingly contradictory signals (high congestion but no host throttling).

\subsection{Evaluation: The OpenRCA Benchmark}
To quantify Loom's efficacy, we evaluated it on the OpenRCA benchmark \citep{xu2025openrca}, comprising three datasets (Telecom, Bank, and Market), where we report the two Market cloudbed instances separately as Market-1 and Market-2. We adapt Loom's synthesis step to output structured JSON matching the benchmark's strict schema. Loom is deterministic by construction, so all reported numbers are exact, contrasting with the \texttt{RCA-Agent} baseline\footnote{\url{https://github.com/microsoft/OpenRCA}}, whose sampled trajectories vary.

\begin{table*}[t]
\centering
\footnotesize
\setlength{\tabcolsep}{4pt}
\begin{tabular}{@{}l|cc|cc|c@{}}
\toprule
\textbf{Dataset} &
  \multicolumn{2}{c|}{\textbf{RCA-Agent + Claude 4.6}} &
  \multicolumn{2}{c|}{\begin{tabular}[t]{@{}c@{}}\textbf{Loom + Claude 4.6}\\[-1pt]\textbf{(Zero-Shot)}\end{tabular}} &
  \begin{tabular}[t]{@{}c@{}}\textbf{Oracle}\\[-1pt]\textbf{(Loom Slate)}\end{tabular} \\
& \textbf{Strict Acc.} & \textbf{Partial Acc.} & \textbf{Strict Acc.} & \textbf{Partial Acc.} & \textbf{Strict Acc.} \\
\midrule
Bank (136) & \textbf{40.44\%} & 49.15\% & 38.97\% & \textbf{51.22\%} & N/A \\
Market-2 (78) & \textbf{35.90\%} & 50.76\% & \textbf{35.90\%} & \textbf{50.85\%} & 70.51\% \\
Market-1 (70) & \textbf{40.00\%} & \textbf{54.41\%} & 28.57\% & 44.53\% & 70.00\% \\
Telecom (51) & \textbf{41.18\%} & \textbf{52.96\%} & 29.41\% & 42.49\% & 35.29\% \\
\midrule
\textbf{Efficiency} &
  \multicolumn{2}{c|}{\begin{tabular}{@{}c@{}}$\sim$62 LLM calls / $\sim$567s\\[-1pt]per incident\end{tabular}} &
  \multicolumn{2}{c|}{\textbf{\begin{tabular}{@{}c@{}}1 LLM call / $\sim$22s per incident\\[-1pt]($\sim$26$\times$ speedup)\end{tabular}}} &
  N/A \\
\bottomrule
\end{tabular}
\caption{Performance comparison on the OpenRCA benchmark (four dataset instances). Loom occupies an accuracy--efficiency Pareto frontier: it matches the agent on Bank and Market-2 and trails on Market-1 and Telecom, while using one LLM call per incident on every dataset ($\sim$26$\times$ speedup). On Market-1 and Telecom, the single-shot LLM struggles to disambiguate dense noise, though the Oracle score shows that consensus often surfaces the correct candidate.}
\label{tab:openrca_all}
\end{table*}

\begin{table*}[!htb]
\centering
\footnotesize
\setlength{\tabcolsep}{4pt}
\begin{tabular}{@{}lcccc@{}}
\toprule
\textbf{Metric} & \textbf{RCA-Agent + Claude 3.5} & \textbf{RCA-Agent + Claude 4.6} & \textbf{Loom + Claude 4.6} & \textbf{Loom + Llama-3.1-8B} \\
\midrule
Easy strict & 17.00\% & 41.94\% & \textbf{46.77\%} & \textbf{46.77\%} \\
Middle strict & 9.00\% & \textbf{42.11\%} & 29.82\% & 26.32\% \\
Hard strict & 0.00\% & 29.41\% & \textbf{41.18\%} & 23.53\% \\
\textbf{Overall strict} & 11.34\% & \textbf{40.44\%} & 38.97\% & 35.29\% \\
Overall partial & 17.00\% & 49.15\% & \textbf{51.22\%} & 46.68\% \\
Avg time / item & N/A & 566.8 s & $\sim$22 s & \textbf{$\sim$17 s} \\
Judge time / item & N/A & $\sim$566 s & $\sim$3.5 s & \textbf{$\sim$1.5 s} \\
\textbf{Speedup vs agent} & N/A & 1$\times$ & $\sim$26$\times$ & \textbf{$\sim$33$\times$} \\
LLM calls / item & 30 to 50 & 36 to 78 ($\sim$62 avg) & 1 & \textbf{1} \\
Model size (judge) & N/A & $\sim$100B+ & $\sim$100B+ & \textbf{8B} \\
\bottomrule
\end{tabular}
\caption{Detailed performance comparison on the OpenRCA Bank benchmark. On this dataset Loom is near the agent's accuracy while delivering a $\sim$26$\times$ speedup with the Claude 4.6 judge and $\sim$33$\times$ with the Llama-3.1-8B judge. The Llama-3.1-8B judge ties the Claude 4.6 judge on easy queries, demonstrating that the consensus stage carries the heavy lifting.}
\label{tab:bank_detailed}
\end{table*}

\subsection{Lessons Learned and Ablation Studies}

Deploying Loom yielded several insights into the trade-offs between quality, efficiency, and system design in real-world settings. To isolate the contribution of each component, we ran a controlled ablation on the OpenRCA Bank benchmark (Table~\ref{tab:ablations}).

\paragraph{Lesson 1: The Cost-Accuracy Pareto Frontier.} 
As shown in Table~\ref{tab:openrca_all}, Loom occupies an accuracy--efficiency Pareto frontier. On Bank and Market-2 it matches the agent's strict accuracy on Market-2 ($35.90\%$) and exceeds its partial accuracy on Bank ($51.22\%$ vs.\ $49.15\%$); it trails on Market-1 and Telecom. On every OpenRCA dataset, \texttt{RCA-Agent} requires $\sim$62 iterative LLM calls and nearly 10 minutes per incident, whereas Loom requires 1 call and $\sim$22 seconds (a $\sim$26$\times$ speedup). Furthermore, Table~\ref{tab:bank_detailed} shows Loom's Bank advantage is most pronounced at the extremes of query difficulty (easy: $46.77\%$ vs.\ $41.94\%$; hard: $41.18\%$ vs.\ $29.41\%$); the agent retains its lead only on two-element queries.

\paragraph{Lesson 2: Decoupling Consensus from LLM Scale.} 
Table~\ref{tab:bank_detailed} demonstrates that Loom's performance is largely independent of LLM scale. Replacing Claude 4.6 with Llama-3.1-8B drops overall strict accuracy by only $\sim$3.7~pp; on single-element queries the two judges are tied. Because the reweighting algorithm handles conflict resolution, the LLM only performs basic summarization, enabling local, small-parameter deployment, a critical requirement for air-gapped or cost-sensitive industrial environments.

\begin{table}[t]
\centering
\footnotesize
\setlength{\tabcolsep}{4pt}
\begin{tabular}{@{}lcc@{}}
\toprule
\textbf{Configuration} & \textbf{Strict Acc.} & \textbf{Partial Acc.} \\
\midrule
Full Loom System & 38.97\% & 51.22\% \\
\midrule
w/o Iter. Reweighting & 33.09\% & 42.70\% \\
\textbf{w/o Redundancy Det.} & \textbf{44.85\%} & \textbf{53.36\%} \\
w/o Both (Raw LLM Synth.) & 35.29\% & 44.49\% \\
\bottomrule
\end{tabular}
\caption{Ablation on the 136-incident OpenRCA Bank benchmark. Iterative centroid reweighting contributes a clear $+5.88$ pp strict-accuracy gain over the no-reweight baseline.}
\label{tab:ablations}
\end{table}

\paragraph{Lesson 3: Negative Results with Static Redundancy.} 
Contrary to our initial hypothesis, removing the static DS-docstring redundancy step actually \textit{improves} accuracy (coincidentally also $+5.88$ pp strict) on the Bank benchmark (Table~\ref{tab:ablations}). Post-hoc diagnosis clarifies the mechanism: in environments with a small, expert-curated DS catalog, docstring-level grouping is too coarse a proxy for true redundancy. It compresses entire reason-families out of the top-$K$ candidate list, making the LLM judge less able to recognize the correct failure category. This highlights a practical lesson: redundancy terms must be dynamically conditioned on outputs rather than statically bound to docstrings when operating over curated rule sets.

\paragraph{Lesson 4: The Oracle and the Limits of Single-Shot Synthesis.}
On Market-1 and Telecom, Loom underperforms the iterative agent. However, the Oracle scores on Market-1 ($70.00\%$) and Market-2 ($70.51\%$) show the continuous consensus algorithm reliably places the true root cause in the top candidates. The remaining gap is the single-shot LLM's inability to disambiguate closely related fault reasons without iterative tool calls. This suggests a hybrid approach (using Loom to rapidly surface candidates and a lightweight agent to disambiguate) could close the accuracy gap while preserving efficiency.

\section{Conclusion}

Loom is a deployable generative consensus framework that aggregates templated, episode-specific diagnostic hypotheses in a continuous embedding space. By shifting conflict resolution from iterative LLM loops to mathematical reweighting, Loom achieves a $\sim$26--33$\times$ inference speedup on OpenRCA and enables the use of small (8B) local models, meeting the strict latency and cost constraints of industrial operations. While single-shot synthesis has limits on highly complex datasets, Loom's deterministic, auditable pipeline fosters SME trust and establishes a practical blueprint for real-world automated diagnostics.

\section{Limitations}
\label{sec:limitations}

While Loom offers substantial improvements in scalability and inference latency, evaluating it against generic, bottom-up agentic solutions reveals several strategic trade-offs and engineering considerations.

\paragraph{Accuracy Gap on Market-1 and Telecom:}
We want to be explicit about an empirical limitation: on two of the four reported OpenRCA instances, Loom is materially less accurate than the iterative \texttt{RCA-Agent} baseline. On Market-1 the gap is $11.43$ pp strict ($28.57\%$ vs.\ $40.00\%$) and on Telecom it is $11.77$ pp strict ($29.41\%$ vs.\ $41.18\%$). The Oracle analysis of Table~\ref{tab:openrca_all} suggests the continuous consensus stage surfaces the correct candidate on Market-1 ($70.00\%$ Oracle), but on Telecom the Oracle itself ($35.29\%$) underperforms the agent, indicating the bottleneck on Telecom is partially in the DS catalog's coverage rather than purely in the single-shot synthesis step. We therefore do not claim parity with iterative agents in general; Loom is competitive at the two extremes of query complexity on Bank and on Market-2, and is meaningfully weaker on the other two benchmarks.

\paragraph{Semantic Coverage and Predictability:} 
A primary limitation of Loom is its reliance on a predefined catalog of Diagnostic Strands. Unlike bottom-up agents that theoretically possess unbounded coverage and attempt to solve novel issues on-the-fly, Loom's coverage is explicitly bounded by its DS catalog. However, this limitation yields high predictability: system operators know exactly what the system can and cannot solve, allowing them to systematically close ``blind spots'' rather than relying on the inconsistent success rates of exploratory agents.

\paragraph{The Cold Start Challenge and Governance:} 
When confronted with a novel ``Black Swan'' failure, an exploratory agent will attempt a zero-shot diagnosis, which risks unverified or hallucinated executive decisions during a crisis. Loom, by contrast, faces a cold start challenge: it requires a human expert or an offline LLM extraction pipeline to define and validate a new DS before it can diagnose the novel issue. While this delays immediate zero-shot resolution of unprecedented failures, it enforces strict governance.

\paragraph{Dependency on Embedding Expressiveness:} 
Because Loom's aggregation relies on vector-space geometry, the system's accuracy is highly sensitive to the nuances of textual embeddings. If the embedding model fails to capture the semantic distance between subtle technical nuances, the iterative reweighting algorithm may conflate distinct root causes. Conversely, bottom-up agents rely deeply on foundational LLMs, which suffer from contextual conflation and may treat distinct hardware errors as having the same semantic representation. Loom's mathematical aggregation enforces strict geometric consistency, provided the underlying embedding space is sufficiently expressive. Furthermore, the technical debt associated with maintaining and updating embedding models is orders of magnitude lower than that of managing monolithic foundational LLMs.

\paragraph{The Limits of Single-Shot Synthesis and Judge Overfocus:} 
While Loom's single-shot LLM synthesis drastically reduces inference latency, our evaluation on the Telecom and Market benchmarks reveals the boundaries of single-pass reasoning. On the Telecom dataset, we observed a phenomenon we term \textit{Judge Overfocus}: when presented with multiple high-confidence candidates featuring saturating anomaly scores across different nodes, the single-shot LLM struggles to weigh competing evidence and often commits to an incorrect, albeit highly anomalous, fault type. Similarly, on the Market-1 benchmark, the single-shot synthesizer exhibited ``same-family reason confusion,'' frequently failing to disambiguate closely related root causes. Crucially, oracle evaluations on the candidate slates showed that the correct hypothesis was present over 70\% of the time, indicating that the continuous consensus stage successfully surfaced the truth, but the single-shot reasoning head lacked the investigative depth to perform final disambiguation.

\paragraph{Reproducibility:}
The algorithmic framework, the iterative reweighting algorithm (Algorithm~\ref{alg:reweighting}), the LLM synthesis prompt (excerpted in Appendix~\ref{app:synthesis_prompt}), and the DS-extraction methodology are specified in this paper and the appendices, so Loom is reproducible in principle on any new domain. Our OpenRCA experiments use the publicly released benchmark of \citet{xu2025openrca}, and the Claude~4.6 and Llama-3.1-8B synthesizers are off-the-shelf API and open-weight models, respectively. Combined with the determinism of the pipeline at temperature 0, this means that an independent re-implementation should obtain numerically identical strict/partial-accuracy figures on OpenRCA up to backend LLM-API nondeterminism.

\bibliography{custom}

@inproceedings{jiang2024megascale,
author = {Jiang, Ziheng and Lin, Haibin and Zhong, Yinmin and Huang, Qi and Chen, Yangrui and Zhang, Zhi and Peng, Yanghua and Li, Xiang and Xie, Cong and Nong, Shibiao and Jia, Yulu and He, Sun and Chen, Hongmin and Bai, Zhihao and Hou, Qi and Yan, Shipeng and Zhou, Ding and Sheng, Yiyao and Jiang, Zhuo and Xu, Haohan and Wei, Haoran and Zhang, Zhang and Nie, Pengfei and Zou, Leqi and Zhao, Sida and Xiang, Liang and Liu, Zherui and Li, Zhe and Jia, Xiaoying and Ye, Jianxi and Jin, Xin and Liu, Xin},
title = {MegaScale: scaling large language model training to more than 10,000 GPUs},
year = {2024},
isbn = {978-1-939133-39-7},
publisher = {USENIX Association},
address = {USA},
booktitle = {Proceedings of the 21st USENIX Symposium on Networked Systems Design and Implementation},
articleno = {41},
numpages = {16},
location = {Santa Clara, CA, USA},
series = {NSDI'24}
}

@inproceedings{jiang2025l4,
author = {Jiang, Zhihan and Huang, Junjie and Yu, Guangba and Chen, Zhuangbin and Li, Yichen and Zhong, Renyi and Feng, Cong and Yang, Yongqiang and Yang, Zengyin and Lyu, Michael},
title = {L4: Diagnosing Large-scale LLM Training Failures via Automated Log Analysis},
year = {2025},
isbn = {9798400712760},
publisher = {Association for Computing Machinery},
address = {New York, NY, USA},
url = {https://doi.org/10.1145/3696630.3728531},
doi = {10.1145/3696630.3728531},
booktitle = {Proceedings of the 33rd ACM International Conference on the Foundations of Software Engineering},
pages = {51–63},
numpages = {13},
location = {Clarion Hotel Trondheim, Trondheim, Norway},
series = {FSE Companion '25}
}

@inproceedings{ratner2016data,
  title={Data programming: Creating large training sets, quickly},
  author={Ratner, Alexander J and De Sa, Christopher M and Wu, Sen and Selsam, Daniel and R{\'e}, Christopher},
  booktitle={Advances in neural information processing systems},
  volume={29},
  year={2016}
}

@article{ratner2017snorkel,
  title={Snorkel: Rapid training data creation with weak supervision},
  author={Ratner, Alexander and Bach, Stephen H and Ehrenberg, Henry and Fries, Jason and Wu, Sen and R{\'e}, Christopher},
  journal={Proceedings of the VLDB Endowment},
  volume={11},
  number={3},
  pages={269--282},
  year={2017},
  publisher={VLDB Endowment}
}

@inproceedings{xu2025openrca,
title={OpenRCA: Can Large Language Models Locate the Root Cause of Software Failures?},
author={Xu, Junjielong and Zhang, Qinan and Zhong, Zhiqing and He, Shilin and Zhang, Chaoyun and Lin, Qingwei and Pei, Dan and He, Pinjia and Zhang, Dongmei and Zhang, Qi},
booktitle={The Thirteenth International Conference on Learning Representations},
year={2025},
url={https://openreview.net/forum?id=M4qNIzQYpd}
}

@inproceedings{fu2020fast,
  title={Fast and three-rious: Speeding up weak supervision with triplet methods},
  author={Fu, Daniel Y and Chen, Mayee F and Sala, Frederic and Hooper, Sarah M and Fatahalian, Kayvon and R{\'e}, Christopher},
  booktitle={International Conference on Machine Learning},
  pages={3280--3291},
  year={2020},
  organization={PMLR}
}

@inproceedings{varma2019learning,
  title={Learning dependency structures for weak supervision models},
  author={Varma, Paroma and Sala, Frederic and He, Ann and Ratner, Alexander and R{\'e}, Christopher},
  booktitle={International Conference on Machine Learning},
  pages={6418--6427},
  year={2019},
  organization={PMLR}
}

@inproceedings{yao2023react,
  title={ReAct: Synergizing Reasoning and Acting in Language Models},
  author={Yao, Shunyu and Zhao, Jeffrey and Yu, Dian and Du, Nan and Shafran, Izhak and Narasimhan, Karthik and Cao, Yuan},
  booktitle={International Conference on Learning Representations (ICLR)},
  year={2023}
}

@inproceedings{wang2023selfconsistency,
  title={Self-Consistency Improves Chain of Thought Reasoning in Language Models},
  author={Wang, Xuezhi and Wei, Jason and Schuurmans, Dale and Le, Quoc and Chi, Ed and Narang, Sharan and Chowdhery, Aakanksha and Zhou, Denny},
  booktitle={International Conference on Learning Representations},
  year={2023}
}

@inproceedings{du2024improving,
author = {Du, Yilun and Li, Shuang and Torralba, Antonio and Tenenbaum, Joshua B. and Mordatch, Igor},
title = {Improving factuality and reasoning in language models through multiagent debate},
year = {2024},
publisher = {JMLR.org},
booktitle = {Proceedings of the 41st International Conference on Machine Learning},
articleno = {467},
numpages = {31},
location = {Vienna, Austria},
series = {ICML'24}
}

@inproceedings{liang2024encouraging,
  title={Encouraging Divergent Thinking in Large Language Models through Multi-Agent Debate},
  author={Liang, Tian and He, Zhiwei and Jiao, Wenxiang and Wang, Xing and Wang, Yan and Wang, Rui and Yang, Yujiu and Shi, Shuming and Tu, Zhaopeng},
  booktitle={Proceedings of the 2024 Conference on Empirical Methods in Natural Language Processing},
  year={2024}
}

@inproceedings{lewis2020rag,
  title     = {Retrieval-Augmented Generation for Knowledge-Intensive {NLP} Tasks},
  author    = {Lewis, Patrick and Perez, Ethan and Piktus, Aleksandra and Petroni, Fabio and Karpukhin, Vladimir and Goyal, Naman and K{\"u}ttler, Heinrich and Lewis, Mike and Yih, Wen-tau and Rockt{\"a}schel, Tim and Riedel, Sebastian and Kiela, Douwe},
  booktitle = {Advances in Neural Information Processing Systems},
  volume    = {33},
  pages     = {9459--9474},
  year      = {2020},
  url       = {https://proceedings.neurips.cc/paper/2020/hash/6b493230205f780e1bc26945df7481e5-Abstract.html}
}

@inproceedings{izacard2021leveraging,
  title     = {Leveraging Passage Retrieval with Generative Models for Open Domain Question Answering},
  author    = {Izacard, Gautier and Grave, Edouard},
  booktitle = {Proceedings of the 16th Conference of the European Chapter of the Association for Computational Linguistics: Main Volume},
  pages     = {874--880},
  year      = {2021},
  month     = apr,
  address   = {Online},
  publisher = {Association for Computational Linguistics},
  url       = {https://aclanthology.org/2021.eacl-main.74/},
  doi       = {10.18653/v1/2021.eacl-main.74}
}

@inproceedings{asai2024selfrag,
  title     = {{Self-RAG}: Learning to Retrieve, Generate, and Critique through Self-Reflection},
  author    = {Asai, Akari and Wu, Zeqiu and Wang, Yizhong and Sil, Avirup and Hajishirzi, Hannaneh},
  booktitle = {The Twelfth International Conference on Learning Representations},
  year      = {2024},
  url       = {https://openreview.net/forum?id=hSyW5go0v8}
}

@inproceedings{zheng2023judging,
  title     = {Judging {LLM-as-a-Judge} with {MT-Bench} and {Chatbot Arena}},
  author    = {Zheng, Lianmin and Chiang, Wei-Lin and Sheng, Ying and Zhuang, Siyuan and Wu, Zhanghao and Zhuang, Yonghao and Lin, Zi and Li, Zhuohan and Li, Dacheng and Xing, Eric P. and Zhang, Hao and Gonzalez, Joseph E. and Stoica, Ion},
  booktitle = {Advances in Neural Information Processing Systems 36: Annual Conference on Neural Information Processing Systems 2023, Track on Datasets and Benchmarks},
  year      = {2023},
  url       = {https://proceedings.neurips.cc/paper_files/paper/2023/hash/91f18a1287b398d378ef22505bf41832-Abstract-Datasets_and_Benchmarks.html}
}

@misc{verga2024replacing,
  title         = {Replacing Judges with Juries: Evaluating {LLM} Generations with a Panel of Diverse Models},
  author        = {Verga, Pat and Hofst{\"a}tter, Sebastian and Althammer, Sophia and Su, Yixuan and Piktus, Aleksandra and Arkhangorodsky, Arkady and Xu, Minjie and White, Naomi and Lewis, Patrick},
  year          = {2024},
  eprint        = {2404.18796},
  archivePrefix = {arXiv},
  primaryClass  = {cs.CL},
  url           = {https://arxiv.org/abs/2404.18796}
}

@inproceedings{jiang2023llmblender,
  title     = {{LLM-Blender}: Ensembling Large Language Models with Pairwise Ranking and Generative Fusion},
  author    = {Jiang, Dongfu and Ren, Xiang and Lin, Bill Yuchen},
  booktitle = {Proceedings of the 61st Annual Meeting of the Association for Computational Linguistics (Volume 1: Long Papers)},
  pages     = {14165--14178},
  year      = {2023},
  publisher = {Association for Computational Linguistics},
  url       = {https://aclanthology.org/2023.acl-long.792/},
  doi       = {10.18653/v1/2023.acl-long.792}
}

\appendix

\section{Example Diagnostic Strand}
\label{app:ds_example}

Figure~\ref{fig:ds_example_code} provides a concrete example of a Diagnostic Strand (DS) implemented in Python. This heuristic diagnoses non-fatal accelerator-fabric link failures. It cross-references unstructured text (syslog signatures) with structured fabric counters and fills a template with episode-specific hosts and counts.

\begin{figure*}[!ht]
\centering
\footnotesize
\begin{verbatim}
@diagnostic_strand(
    name="ds_remap_nonfatal_link_failure",
    source="incident_db",
    reliability="high",
)
def ds_remap_nonfatal_link_failure(ctx: DatacenterContext):
    """REMAP non-fatal fabric error: root cause: link physical-layer
    training failure, likely a faulty transceiver, cable, or port instability."""
    if ctx.syslog.empty or "message" not in ctx.syslog.columns:
        return ABSTAIN
    hits = ctx.syslog[ctx.syslog["message"].str.contains(
        r"FabricFault\(PCI:[^)]+\):\s*ERR_LINK_REMAP_NF",
        case=False, regex=True, na=False,
    )]
    if hits.empty:
        return ABSTAIN

    hosts = sorted(hits["hostname"].unique())
    details = []
    if "rx_error_count" in ctx.fabric.columns:
        rx = ctx.fabric["rx_error_count"].fillna(0)
        if (rx > 0).any():
            details.append(f"elevated link rx errors (max={rx.max():.0f})")
    if "link_retry_codes" in ctx.fabric.columns:
        rt = ctx.fabric["link_retry_codes"].fillna(0)
        if (rt > 0).any():
            details.append(f"concurrent link-layer retransmissions (max={rt.max():.0f})")

    desc = (f"{len(hits)} REMAP non-fatal events on {', '.join(hosts)}: "
            "fabric link physical-layer training failure "
            "(likely faulty transceiver / cable / port instability)")
    if details:
        desc += " [" + "; ".join(details) + "]"
    return desc
\end{verbatim}
\caption{Example Diagnostic Strand (DS) for the datacenter domain. The strand cross-references a syslog regex signature with two structured fabric-telemetry counters via a unified \texttt{DatacenterContext}, fills an episode-specific template when its conditions fire, and otherwise returns \texttt{ABSTAIN}.}
\label{fig:ds_example_code}
\end{figure*}

\section{Synthesis System Prompt}
\label{app:synthesis_prompt}

Figure~\ref{fig:synthesis_prompt_text} shows an excerpt from the system prompt used to guide the final LLM synthesis stage. The prompt is strictly constrained to prevent the LLM from acting as an independent investigator. Instead, it forces the model to reason exclusively over the ordered, mathematically denoised evidence provided by the continuous consensus algorithm, ensuring the final narrative remains grounded in the raw telemetry.

\begin{figure*}[!ht]
\centering
\scriptsize
\begin{verbatim}
You are an expert infrastructure troubleshooting engineer. 
You will receive a set of diagnostic observations about a 
failed job, listed in order of decreasing importance. 
Your job is to synthesize them into a single, coherent 
root cause explanation for the operations team.

Rules:
- Earlier observations are more important, give them 
  more prominence.
- When multiple observations describe different facets 
  of the same failure cascade, connect them causally.
- Preserve specific technical details (error codes, 
  hostnames, metric values) verbatim from the input.
- Do NOT invent facts, metrics, or details beyond what 
  the input observations report.
- Do NOT reference internal diagnostic function names, 
  weights, or scoring mechanics.
\end{verbatim}
\caption{Excerpt from the Loom LLM synthesis system prompt. The model is constrained to reason over the ordered DS outputs, ensuring the final narrative is causally sound and strictly grounded in the telemetry-backed heuristics.}
\label{fig:synthesis_prompt_text}
\end{figure*}

\section{Offline Learning Pipeline}
\label{app:offline_learning}

\begin{figure*}[!ht]
\centering
\includegraphics[width=\textwidth]{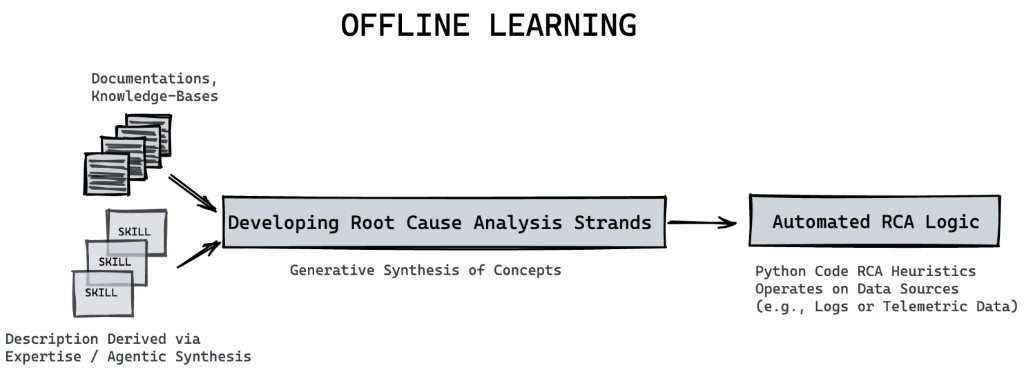}
\caption{The offline learning process for generating Diagnostic Strands. An LLM agent ingests diverse sources of information, such as system documentation and Subject Matter Expert (SME) skills, to automatically extract and generate programmatic RCA heuristics (DSs) that operate on structured data sources.}
\label{fig:offline_learning_diagram}
\end{figure*}

Figure~\ref{fig:offline_learning_diagram} illustrates the offline learning pipeline used to construct the catalog of Diagnostic Strands. Generative AI agents process unstructured knowledge bases (e.g., system documentation, incident tickets) and extract them into discrete, executable Python functions. This pre-compilation step prevents the system from having to blindly diagnose massive telemetry dumps at inference time.

\section{InfiniBand Switch Environmental Factor}
\label{app:ib_switch}

Large GPU clusters interconnect nodes with InfiniBand (IB), a high-bandwidth switched fabric used as the datacenter compute network. An IB switch exposes tens of ports, each terminating a cable or transceiver toward a host or another switch. Isolated port failures (a single bad cable or optic) are common; correlated degradation of many ports on the \emph{same} switch is not.

The Diagnostic Strand \texttt{ds\_ibjf\_switch\_environmental} encodes that spatial rule. It groups IB telemetry by switch and marks a port degraded if its effective bit-error rate exceeds $10^{-12}$ or its packet-loss-retry ratio is at least $10^{-2}$. The strand fires only when a \emph{majority} of the switch's active ports are degraded ($>50\%$ and at least three ports). Independent per-port faults are then unlikely; a chassis-level environmental cause (cooling, airflow, dust, heat) is the appropriate hypothesis. Temperature is the high-probability corroboration rather than an automatic label: if device temperature is at least $80^\circ$C, module temperature at least $70^\circ$C, or thermal opcodes/flags are set, the filled template names an environmental/thermal factor; otherwise it still reports a common-mode environmental factor and states that heat versus dust, smoke, or airflow is ungrounded without thermal evidence. Figure~\ref{fig:ib_switch_code} shows a condensed listing of this decision rule.

\begin{figure*}[!ht]
\centering
\footnotesize
\begin{verbatim}
def ds_ibjf_switch_environmental(ctx):
    """Majority port degradation on one IB switch
    (environmental / thermal common-mode factor)."""
    for switch_id, ports in group_by_switch(ctx.ib):
        active = [p for p in ports if is_active(p)]
        bad = [p for p in active if (
            p.effective_ber > 1e-12
            or p.plr_retry_ratio >= 1e-2
        )]
        if len(bad) < 3 or len(bad) / len(active) <= 0.5:
            continue  # abstain on this switch
        if has_thermal_corroboration(active):
            # device>=80C, module>=70C, or thermal flags
            return (
                f"Hypothesis: Environmental / thermal factor "
                f"on switch {switch_id}: majority degraded "
                f"{len(bad)}/{len(active)} active ports; "
                f"remediate cooling / environment"
            )
        return (
            f"Hypothesis: Common-mode environmental factor "
            f"on switch {switch_id}: majority degraded "
            f"{len(bad)}/{len(active)} active ports; cause "
            f"ungrounded (heat vs dust/smoke/airflow)"
        )
    return ABSTAIN
\end{verbatim}
\caption{Condensed Diagnostic Strand for correlated InfiniBand port degradation on a single switch. Independent random cable/optic faults rarely hit a majority of a chassis; the template is populated with the switch identity and the affected/active port counts, and names heat only when thermal telemetry corroborates.}
\label{fig:ib_switch_code}
\end{figure*}

\section{OpenRCA Diagnostic Strand Catalog}
\label{app:catalog}

Table~\ref{tab:catalog_stats} reports the OpenRCA Diagnostic Strand catalogs used in our evaluation. Market-1 and Market-2 share a single Market catalog. A strand function may instantiate many entity-, time-, and metric-specific hypotheses on one incident, hence Bank's mean of 340.93 raw hypotheses versus 7.92 fired functions. Only the ranked top-$K$ hypotheses are passed to synthesis.

\begin{table}[h]
\centering
\footnotesize
\setlength{\tabcolsep}{4pt}
\begin{tabular}{@{}lrrrr@{}}
\toprule
\textbf{Dataset} & \textbf{Inc.} & \textbf{DS fns.} & \textbf{Mean fired} & \textbf{Mean hyps.} \\
\midrule
Bank & 136 & 19 & 7.92 & 340.93 \\
Telecom & 51 & 17 & 3.94 & 13.25 \\
Market-1 & 70 & 21$^*$ & 7.29 & 13.14 \\
Market-2 & 78 & 21$^*$ & 7.22 & 15.69 \\
\bottomrule
\end{tabular}
\caption{OpenRCA Diagnostic Strand catalogs. $^*$Market-1 and Market-2 share one catalog. Catalog construction required about one engineer-week per dataset (schema integration, authoring, testing, and evaluation).}
\label{tab:catalog_stats}
\end{table}

\section{Worked OpenRCA Example}
\label{app:worked_example}

We reconstruct one submitted Bank inference from evaluation traces (query 82; window 2021-03-09 19:30--20:00 UTC+8). The query asks for the root-cause reason of a single failure. Ground truth is \texttt{Tomcat03} / high CPU usage / 19:42; official strict score is $1.0$.

\paragraph{Input rows.} Container CPU utilization on \texttt{Tomcat03} is idle through 19:41 ($\approx$26.6), jumps at 19:43 (82.7), saturates at 19:44--19:47 (99.1--99.8), and recovers by 19:49 (26.2). The same window contains GC allocation-failure log lines and Tomcat03 trace spans (e.g., duration 44\,ms).

\paragraph{Firing and abstaining strands.} Of 19 Bank strand functions, 11 fire and produce 206 instantiated hypotheses (192 from a low-reliability fan-out that tentatively assigns many reason labels to stressed pods). Direct firings include sustained CPU on \texttt{Tomcat03}, a metric+trace composite that names the same CPU conclusion at high reliability, disk/JVM/network distractors on neighboring pods, two trace-concentration strands, and a GC-log strand. The remaining eight functions abstain (memory, packet loss, JVM CPU load, slow query, TCP, mrt-memory, and two JVM composites). A representative filled template is:

\begin{quote}
\footnotesize
Root cause: sustained CPU utilization anomaly on Tomcat03 (4 consecutive minutes). component=Tomcat03; reason=high CPU usage; time\_hint=2021-03-09 19:43:00; reliability=medium. Evidence: KPI CPU utilization, high-direction peak $z=236.67$ over 4 minutes.
\end{quote}

\paragraph{Ranked synthesis inputs.} After embedding-centroid reweighting, collapse to one hypothesis per (component, reason) and retain the top $K{=}8$:
\begin{enumerate}
\footnotesize
\item $0.898$: Tomcat03, high CPU usage, 19:43 (CPU strand)
\item $0.878$: Tomcat03, network latency, 19:43 (fan-out)
\item $0.869$: Tomcat03, high disk space usage, 19:43 (fan-out)
\item $0.865$: Tomcat02, network latency, 19:43 (fan-out)
\item $0.862$: Tomcat03, network packet loss, 19:43 (fan-out)
\item $0.859$: Tomcat03, high memory usage, 19:43 (fan-out)
\item $0.859$: Tomcat03, high disk I/O read usage, 19:43 (fan-out)
\item $0.852$: Tomcat03, JVM OOM Heap, 19:43 (fan-out)
\end{enumerate}

The true pair is rank 1. The synthesizer then emits OpenRCA JSON (score $1.0$):
{\scriptsize
\begin{verbatim}
{"1": {
  "root cause occurrence datetime":
      "2021-03-09 19:42:00",
  "root cause component": "Tomcat03",
  "root cause reason": "high CPU usage"}}
\end{verbatim}
}

The CPU strand's \texttt{time\_hint} is 19:43 (first sustained minute on the one-minute KPI grid); the synthesizer reports 19:42, matching the record label.

\end{document}